\documentclass{article}

\usepackage{arxiv} 

\usepackage[utf8]{inputenc} 
\usepackage[T1]{fontenc}    
\usepackage{hyperref}       
\usepackage{url}            
\usepackage{booktabs}       
\usepackage{amsfonts}       
\usepackage{nicefrac}       
\usepackage{microtype}      
\usepackage{lipsum}

\usepackage{tikz,pgf}
\usetikzlibrary{shapes.geometric,arrows,decorations.pathmorphing, backgrounds, positioning, fit, petri}
\usepackage{adjustbox}

\definecolor{light-grey}{gray}{0.85}

\usepackage{algorithm}
\usepackage{algpseudocode}
\usepackage{graphicx}	
\usepackage{cancel}
\usepackage{multirow}

\usepackage[font=small,labelfont=bf,tableposition=top]{caption}
\DeclareCaptionLabelFormat{andtable}{#1~#2  \&  \tablename~\thetable}

\title{The MatrixX Solver for Argumentation Frameworks}

\author{
    Maximilian ~Heinrich \\
  Intelligent Systems Department\\
  Computer Science Institute\\
  Leipzig University\\
  \texttt{mheinrich@informatik.uni-leipzig.de} \\
}

\begin{document}
\maketitle
\begin{abstract}
MatrixX is a solver for Abstract Argumentation Frameworks.  Offensive and defensive  properties of an Argumentation Framework are notated in a matrix style. Rows and columns of this matrix are systematically reduced by the solver. This procedure is implemented through the use of hash maps in order to accelerate  calculation time. MatrixX works for stable and complete semantics and was designed for the ICCMA'21 competition.
\end{abstract}

\section{Introduction}
Main idea for the MatrixX solver is the observation that an Argumentation Framework (AF) can be represented as a matrix,  where the offensive and defensive properties of the AF are displayed as rows and columns. The solver shrinks the matrix in a systematic way, which is inspired by Knuth's Algorithm X \cite{Knuth00}. Therefore, matrix approach plus Algorithm X results in the chosen solver name. The solver and instructions for use can be found at:
\begin{center}
\url{https://github.com/kmax-tech/MatrixX}.  
\end{center}
MatrixX supports stable and complete semantics according to ICCMA specifications \cite{LagniezLMR20}. The paper starts with a very short recap about AFs, then the operating principle of the solver is explained via a practical example.

\section{Argumentation Frameworks}
An AF, first introduced by Dung \cite{Dung95}, is a directed graph $F = (A,R)$ with a set of arguments $A$ and an attack relation $R \subseteq A \times A$. Regarding semantics, we state that a set $S \subseteq A$ is conflict-free ($cf$) if $S$ does not attack any of its elements and $S$ is admissible ($adm$) if  $S$ is conflict-free and $S$ defends all its elements against their attackers. A node is defended if all of its attackers are counterattacked. In addition, the set $S$ is complete ($co$) if it is admissible and $S$ contains all elements it defends. The set $S$ is stable ($st$) if it is conflict-free and each argument  which is not in $S$ is getting attacked \cite{GagglLMW20}. For further specification we say that for a node $a \in A$  the set $\{ a\}^{+}$ contains all nodes which $a$ attacks, formally stated as $\{ a\}^{+}  = \{ b | (a,b) \in R \}$. In the paper we will refer to $\{ a\}^{+}$ as the offensive properties of node $a$. Vice versa, with defensive properties $\{a\}^{-}$ we denote the set of all nodes which are attacking $a$, stated as $\{ a\}^{-}  = \{ b | (b,a) \in R \}$. The nodesize of an AF is specified as $n=|A|$, which is the number of arguments in $A$. For a better illustration of the solver mechanics we take Figure \ref{AFE} with $ X = (\{a,b,c,d\},\{(a,b),(b,c),(c,d),(d,a)\})$ as running example, which has the stable interpretations $st(X) = \{ \{a,c\}, \{d,b\} \}$  and complete interpretations $co(X) = \{ \{\}, \{a,c\}, \{d,b\} \}$.

\begin{figure}[ht]
\centering
 \begin{minipage}[c]{0.45\textwidth}
 \centering

 \begin{adjustbox}{width=2.5cm}

\begin{tikzpicture}
  \node[draw=black, minimum size=7mm, circle, thick]  (a) at (0,0) {$a$};
 \node[draw=black, minimum size=7mm, circle, thick] (b) at (2,0) {$b$};
\node[draw=black, minimum size=7mm, circle, thick] (d) at (0,-2) {$d$};
\node[draw=black, minimum size=7mm, circle, thick] (c) at (2,-2) {$c$};
\path(a) edge [->, thick, bend left] (b);
\path(b) edge [->, thick, bend left](c);
\path(c) edge [->, thick, bend left] (d);
\path(d) edge [->, thick, bend left] (a);
\end{tikzpicture}
\end{adjustbox}
 \captionof{figure}{An Example AF $X$}
 \label{AFE}
 \end{minipage}
  \begin{minipage}[c]{0.42\textwidth}
  \centering
\begin{tabular}[b]{c|c|c|c|c|c}
\multicolumn{1}{c}{} & \multicolumn{4}{@{}c}{$\overbrace{\rule{5em}{0pt}}^{DEF}$} \\
\cline{1-5}
 nodes &$a$ &$b$ &$c$&$d$ & \multirow{3}{*}{\rotatebox{270}{$\overbrace{\rule{4em}{0pt}}^{OFF}$}}\\
\cline{1-5}
 $a$ &$0$ &$1$ &$0$ &$0$  \\
$b$ &$0$ &$0$ &$1$ &$0$ \\
$c$ &$0$ &$0$ &$0$ &$1$\\
$d$ &$1$ &$0$ &$0$ &$0$ \\
\cline{1-5}
\end{tabular}
\captionof{table}{Matrix $M_X$}\label{MX}
  \end{minipage}
\caption*{The AF $X$ and its matrix representation $M_X$. If we look  in the matrix e.g. at the row for $a$ we see an entry at $b$, meaning that $a$ attacks $b$ and vice versa that $b$ is attacked by $a$}
 \end{figure}

\section{Matrix Representation of AFs} 

A matrix representation $M_F$ for an AF $F$ with nodesize $n$ is a $n \times n$ matrix, where each row and column represents a corresponding node of the AF. For better orientation we use the nodes directly as indizes for the matrix. If a node attacks another node the entry in the matrix is marked with $1$ otherwise $0$ is used. More formally, we start with a zero matrix and add the entries  $\forall a \in A: b\in \{a\}^{+}  \rightarrow {M_F}_{ab} = 1$. Because offensive  and defensive properties are closely related, we also obtain the defensive properties of all nodes with this construction process. 
If we look at the rows of the created matrix we can see the offensive properties of the corresponding node and if we look at a column, we get the defensive properties of the represented node, every existing relation is marked with $1$. The solver operates on shrinking the offensive and defensive entries in $M_F$, where the rows and columns are removed independently. In order to talk about the removal procedure we introduce the sets $OFF$ and $DEF$, where $OFF$ is representing all rows and $DEF$ the columns of the matrix. If a node is contained in $OFF$ or $DEF$ then the corresponding row respectively column is still considered for the following operations.  In addition, offensive and defensive properties of the nodes are calculating w.r.t. to the existing rows and columns. In this context, for a node $a$, $\{a\}^{+} = \{ b | (a,b) \in R \land b \in DEF \}$ and $\{a\}^{-} = \{ b | (b,a) \in R \land b \in OFF \}$. At the beginning we have $A = OFF = DEF$.  Also we start with the current extension $E = \emptyset$. The set $E$ is the storage for nodes which are obtained during the calculation. For our example Figure \ref{AFE}, which has nodesize $4$, the corresponding matrix representation is stated in Table \ref{MX}.  The stable semantics can now be obtained by repeatedly  applying Algorithm \ref{MatrixX-St}.

\begin{algorithm}
\caption{MatrixX - Iteration for Stable Semantics}\label{MatrixX-St}
\begin{algorithmic}[1]

    \Require Matrix $M$, with information about $OFF,DEF$ and the current extension $E$ 
  \State  $op\_range  =   OFF \cap DEF$ 
    
    \If {$ \exists i \in op\_range$, where $\{i\}^{-}=  \emptyset$} \Comment{a node which is not attacked exists}
    \State select $i$
    \State $C$ = node\_choosen($M$,$i$) \Comment{apply  \texttt{Node\_Chosen} function to copy of $M$}
    \State \Return $L$  = [$C$] \Comment{return matrix $C$}
    \Else
     \State select $i = min(f(op\_range))$, where $f(x) =|\{x\}^{-} |$ \Comment{select node with least attackers}
    \State $C$ = node\_choosen($M$,$i$) \Comment{apply  \texttt{Node\_Chosen} function to copy of $M$}
    \State $N$ = node\_not\_choosen($M$,$i$) \Comment{apply \texttt{Node\_Not\_Chosen} function to copy of $M$}
    \State \Return$L$ = [$C$,$N$]  \Comment{return matrices $C$ and $N$}
    \EndIf
 \State   
\Function{node\_chosen}{matrix $M$, node $i$}
\State $ E = E \cup \{i\} $  \Comment{add node $i$ to extension $E$}
\State $OFF =  OFF \setminus  (i \cup \{i\}^{+} \cup \{i \}^{-} )$  \Comment{erase node $i$ with its offensive and defensive properties}
\State $DEF = DEF \setminus ( i \cup \{i\}^{+} ) $ \Comment{erase node $i$ and its offensive properties}
\State \Return $M$
\EndFunction
 \State
\Function{node\_not\_chosen}{matrix $M$, node $i$}
\State $OFF =  OFF \setminus  \{i \}$  \Comment{erase node $i$}
\State \Return $M$
\EndFunction
     
\end{algorithmic}
\end{algorithm}

The algorithm is applied iteratively for each obtained matrix in $L$. If $OFF = DEF = \emptyset$,  a stable interpretation is found and the current extension $E$ is marked as a valid interpretation. If $OFF$ is empty but not $DEF$, we abandon the current matrix. In the following, we illustrate the procedure with Table \ref{MX} as example. As we start no offensive or defensive information have been erased, resulting in $OFF \cap DEF = A$ and $E = \emptyset$. No node is unattacked. We therefore choose the node with the least number of attackers, due to heuristic reasons. In our case we have multiple options, we decide to use node $d$ and create two copies of $M_X$, further referenced as submatrices.
For the first submatrix we apply the function \texttt{Node\_Chosen}. Hence we add $d$ to our current extension $E$, remove  $\{d,a,c\}$ from $OFF$ and erase  $\{d,a\}$ from $DEF$. The set $OFF$ is representing possible node combinations, which can be chosen from. 
Therefore if $d$ is chosen to be in the extension $E$ we remove it from $OFF$.  Node $d$ attacks $a$ and gets attacked by $c$.  This means that $a$ and $c$ cannot be together with $d$ in the same extension  and are getting removed from $OFF$ too.  $DEF$ is representing the extensionality of the AF. In order to attain stable semantics all nodes must be either in the extension $E$ or get attacked. For that reason only $d$ and $a$ can be removed from $DEF$. Node is $d$ is removed because it is in the extension and $a$ because it is getting attacked. This means that $c$ needs to be eliminated from $DEF$ by another node later in order to obtain a stable interpretation. For the second submatrix we apply the function \texttt{Node\_Not\_Chosen}. For this submatrix we have decided that $d$ shall not be included in the extension $E$, therefore this node is not able to attack any other nodes. Consequently we just remove $d$ from $OFF$. No elements from $DEF$ are getting erased. The two obtained submatrices are shown in Table \ref{chosen} and \ref{notchosen}. We continue to apply Algorithm \ref{MatrixX-St} for the generated submatrices till all stable extensions are obtained.

\begin{table}[!ht]
\centering
 \begin{minipage}[c]{0.45\textwidth}
 \centering
 \begin{tabular}{c|c|c|c|c}
\toprule
 nodes &\xcancel{$a$} &$b$ &$c$&\xcancel{$d$} \\ 
   \midrule
\xcancel{$a$} &\xcancel{$0$} &\xcancel{$1$} &\xcancel{$0$} &\xcancel{$0$} \\
$b$ & \xcancel{$0$} &$0$ &$1$ &\xcancel{$0$} \\
\xcancel{$c$} & \xcancel{$0$} &\xcancel{$0$} &\xcancel{$0$} &\xcancel{$1$} \\
\xcancel{$d$} &\xcancel{$1$} &\xcancel{$0$} &\xcancel{$0$} &\xcancel{$0$} \\
\bottomrule
\end{tabular}
\caption{$M_X$ after  \texttt{Node\_Chosen} function}
\label{chosen}
  \end{minipage}
  \begin{minipage}[c]{0.45\textwidth}
  \centering
\begin{tabular}{c|c|c|c|c}
\toprule
 nodes &$a$ &$b$ &$c$&$d$ \\ 
  \midrule
$a$ &$0$ &$1$ &$0$ &$0$ \\
$b$ &$0$ &$0$ &$1$ &$0$ \\
$c$ &$0$ &$0$ &$0$ &$1$ \\
\xcancel{$d$} &\xcancel{$1$} &\xcancel{$0$} &\xcancel{$0$} &\xcancel{$0$} \\
\bottomrule
\end{tabular}
\caption{$M_X$ after \texttt{Node\_Not\_Chosen} function}
\label{notchosen}
 \end{minipage}
\end{table} 

For complete semantics a slight variation of Algorithm \ref{MatrixX-St} is used. Here the function \texttt{Node\_Not\_Chosen} removes node $i$ from $OFF$, but for all following defensive properties evaluations of other nodes node $i$ is still considered. The complete semantics demands that all attackers must be attacked from the elements in $E$, therefore offensive properties of nodes can only be erased in case a node is selected to be in the extension. In order to obtain the complete interpretations for each generated submatrix the elements in $op\_range$ are checked. If  $op\_range$ contains no unattacked nodes and each node in the current extension $E$ is unattacked, we found a complete interpretation. If the former condition is removed one could also easily calculate the admissible interpretations, though this is not supported by the solver.

\section{Conclusion}
Goal of the solver is that rows and columns of a matrix $M_F$ are erased as fast as possible. With increasing nodesize $n$ the size of the matrix grows quadratically, aggravating the evaluation. In order to do this more efficiently the solver  works with hash maps. Consequently an entry for a node only contains information about its defensive and offensive properties. This way zero entries in the corresponding matrix are omitted. This procedure was also inspired by Knuth's Algorithm X, which uses the concept of dancing links. In addition the solver copies the matrix after each step, therefore backtracking and restoring of previous configurations is not required in order to test all node combinations for an extension. For additional speed-up of the calculation time we first evaluate the grounded extension of the AF, which is used as initial input. Self-attacking nodes are also handled separately. Despite these optimizations, the solver had to fight with time-outs in the competition and performed rather slow. All of its competitors are using  compiled languages such as C++ in some form, showing that MatrixX, which exclusively uses Python as an interpreted language, cannot successfully compete against such approaches. On the other hand MatrixX proposes an interesting approach for solving AFs, which has not been proceeded so  far. Hence it would be interesting to see how fast a compiled version of MatrixX could perform in order to compare solvers which are operating on the same hardware level. Also it seems interesting to modify the described matrix shrinking procedure w.r.t. other semantics.

\bibliographystyle{unsrt}  
\bibliography{references} 
\end{document}